\documentclass[runningheads]{llncs}

\let\llncssubparagraph\subparagraph
\let\subparagraph\paragraph
\usepackage[compact]{titlesec}
\let\subparagraph\llncssubparagraph

\usepackage[utf8]{inputenc}
\usepackage{graphicx,psfrag,epsf}
\usepackage{enumerate}
\usepackage{natbib}

\usepackage{hyperref}
\usepackage{lmodern}
\usepackage[fleqn]{amsmath}
\usepackage{amssymb,amsfonts}
\usepackage{color}
\usepackage{fancyvrb}
\usepackage{caption}
\usepackage{latexsym}
\usepackage{transparent}
\usepackage{pifont}
\usepackage{exscale}
\usepackage{dcolumn}
\usepackage{graphicx}
\usepackage{xcolor}
\usepackage{ragged2e}
\usepackage{longtable}
\usepackage{booktabs}
\usepackage{tabularx}
\usepackage{comment}
\usepackage{float}
\usepackage{dsfont}
\usepackage{csquotes}
\usepackage{rotating}
\usepackage{enumitem}
\usepackage{titlesec}
\usepackage{pgf}
\usepackage{bm}
\usepackage{footnote}
\makesavenoteenv{tabular}
\makesavenoteenv{table}
\usepackage{twoopt}
\usepackage{algorithm}
\usepackage{algpseudocode}
\usepackage{multirow}
\usepackage{makecell}
\usepackage{diagbox}
\usepackage{lscape}
\usepackage{pdflscape}
\usepackage{layouts}
\usepackage{appendix}

\usepackage{chapterbib}

\ifdefined\N                                                                
\renewcommand{\N}{\mathds{N}}                                                
\else
  \newcommand{\N}{\mathds{N}}
\fi
\newcommand{\R}{\mathds{R}}                                                 
\ifdefined\C
  \renewcommand{\C}{\mathds{C}}                                             
\else
  \newcommand{\C}{\mathds{C}}
\fi

\def\argmin{\mathop{\sf arg\,min}}                                          
\newcommand{\fp}[2]{\frac{\partial #1}{\partial #2}}                        

\newcommand{\xb}{\mathbf{x}}													

\newcommand{\D}{\mathcal{D}}                                                
\newcommand{\Dset}{\left( \left(\mathbf{x}^{(1)}, y^{(1)}\right), \ldots, \left(\mathbf{x}^{(n)},  y^{(n)}\right)\right)}    
\newcommand{\yvec}{\left(y^{(1)}, \hdots, y^{(n)}\right)^T}                 
\renewcommand{\xi}[1][i]{\mathbf{x}^{(#1)}}                                          
\newcommand{\yi}[1][i]{y^{(#1)}}                                            
\newcommand{\xivec}{\left(x^{(i)}_1, \ldots, x^{(i)}_p\right)^T}            
\newcommand{\xjvec}{\left(x^{(1)}_j, \ldots, x^{(n)}_j\right)^T}            

\newcommand{\fh}{\hat{f}}                                                   
\newcommand{\fxi}{f\left(\mathbf{x}^{(i)}\right)}                                        

\newcommand{\thetab}{\bm{\theta}}											
\newcommand{\thetabh}{\bm{\hat\theta}}											

\newcommand{\Lxyi}{L\left(\yi, \fxi\right)}                                 

\renewcommand{\ll}{\ell}                                                    

\newcommand{\fmh}{\hat{f}^{[m]}}                                            
\newcommand{\fmdh}{\hat{f}^{[m-1]}}                                         
\newcommand{\rmm}{\tilde{\mathbf{r}}^{[m]}}                                                  
\newcommand{\rmi}{\tilde{r}^{[m](i)}}                                               
\renewcommand{\mathbf}{\bm}
\renewcommand{\rmm}{\bm{r}^{[m]}}
\renewcommand{\rmi}{r^{[m](i)}}
\renewcommand{\ll}{\mathchar"321C}

\newcommand*{\tran}{{\mkern-1.5mu\mathsf{T}}}

\renewcommand{\yvec}{(y^{(1)}, \dots, y^{(n)})^\tran}
\renewcommand{\xivec}{(x^{(i)}_1, \dots, x^{(i)}_p)^\tran}
\renewcommand{\xjvec}{(x^{(1)}_j, \dots, x^{(n)}_j)^\tran}

\begin{document}

\title{Automatic Componentwise Boosting:\\ An Interpretable AutoML System}

\author{Coors Stefan\inst{1}\orcidID{0000-0002-7465-2146} \and
Schalk Daniel\inst{1}\orcidID{0000-0003-0950-1947} \and
Bischl Bernd\inst{1}\orcidID{0000-0001-6002-6980} \and
Rügamer David\inst{1}\orcidID{0000-0002-8772-9202}}
\authorrunning{S. Coors et al.}
%
\institute{Department of Statistics, LMU Munich, Germany\\
\email{\{firstname.lastname\}@stat.uni-muenchen.de}}
\maketitle              

\begin{abstract}

In practice, machine learning (ML) workflows require various different steps, from data preprocessing, missing value imputation, model selection, to model tuning as well as model evaluation. Many of these steps rely on human ML experts. AutoML -- the field of automating these ML pipelines -- tries to help practitioners to apply ML off-the-shelf without any expert knowledge. Most modern AutoML systems like auto-sklearn, H20-AutoML or TPOT aim for high predictive performance, thereby generating ensembles that consist almost exclusively of black-box models. This, in turn, makes the interpretation for the layperson more intricate and adds another layer of opacity for users. We propose an AutoML system that constructs an interpretable additive model that can be fitted using a highly scalable componentwise boosting algorithm. Our system provides tools for easy model interpretation such as visualizing partial effects and pairwise interactions, allows for a straightforward calculation of feature importance, and gives insights into the required model complexity to fit the given task. We introduce the general framework and outline its implementation \texttt{autocompboost}. To demonstrate the frameworks efficacy, we compare \texttt{autocompboost} to other existing systems based on the OpenML AutoML-Benchmark.
Despite its restriction to an interpretable model space, our system is competitive in terms of predictive performance on most data sets while being more user-friendly and transparent.
\end{abstract}

\section{Introduction and Related Work}


Machine learning (ML) models achieve state-of-the-art performances in many different fields of application. Their increasing complexity allows them to be adapted well to non-trivial data generating processes. However, applying ML in practice is usually accompanied with many more hurdles that require both time and expert knowledge. Challenges in the application of ML are, amongst others, proper data preprocessing, missing value imputation, and hyperparameter optimization (HPO). This so-called ML pipeline is usually non-trivial and requires a \enquote{human in the loop}. AutoML systems~\citep{amlb2019} such as Auto-WEKA~\citep{kotthoff2019auto}, Auto-sklearn~\citep{feurer2019auto}, or autoxgboost~\citep{autoxgboost} attempt to automate ML pipelines based on well-defined routines. Human expert knowledge is encoded into an automated process to reduce input from the end user. Automation can also be more time-efficient and superior in predictive performance. To this end, many AutoML systems do not limit themselves to one type of ML model but instead try to solve, e.g., a Combined Algorithm Selection and Hyperparameter Optimization problem \citep[CASH;][]{kotthoff2019auto}. Results of AutoML systems are thus usually large ensembles of different models. Recent frameworks like AutoGluon-Tabular \citep{erickson2020autogluontabular} and Auto-PyTorch Tabular \citep{zimmer2021autopytorch} even incorporate deep neural networks in their systems to further increase the model complexity.

Allowing AutoML systems to build complex ensembles of models increases the likelihood of the automated system to work well on most given data sets without further user input. The complexity of these ensembles, however, makes it harder for practitioners to understand the model's decision process -- an aspect of applied ML that is often at least as important as the predictive performance itself. This results in AutoML systems often not being adopted in practice \citep{drozdal2020trust} and leads researchers in the field of AutoML to raise the question of a trade-off between predictive performance and interpretability \citep{pfisterer2019human,freitas2019interpretability,xanthapoulos2020loop}. In this work, we hypothesize that this model complexity is in many cases unnecessary; an interpretable AutoML system performs equally well on most tasks, yet without the need to use additional interpretation tools or the uncertainty about the model's decisions.  

\subsection{Our Contribution}

We propose a scalable and flexible solution to both automate the ML pipeline targeted by most of the existing AutoML systems and to provide an inherently interpretable model that 1) does not require post-model fitting explanation methods, 
2) automatically yields all the characteristics to understand the final chosen model, while 3) yielding (close to) state-of-the-art performance on most practical use cases.
Our methodological contribution based on a novel stage- and componentwise boosting algorithm is accompanied by an application as well as benchmark experiments to underline the idea and efficacy of our approach. We have implemented our AutoML system in the \texttt{R}~\citep{rmanual} package \texttt{autocompboost} available on GitHub\footnote{\url{github.com/Coorsaa/autocompboost}}. 

\section{Automatic Componentwise Boosting}\label{sec:autocboost}

In the following, we explain our framework in two steps. First, Section~\ref{subsec:cboost} describes the details of our proposed algorithm to fit models based on the pre-processed data. The second part (Section~\ref{subsec:autocboost-parts}) is concerned with the automation of applying the algorithm to a given task.

\subsection{Fitting Engine} \label{subsec:cboost}

Instead of using an ensemble of different models, we propose using componentwise boosting \citep[CWB;][]{buhlmann2003boosting} as a fitting engine. CWB uses additive models as base learners iteratively fitted to pseudo residuals in each boosting iteration with a learning rate $\nu\in\R_+$. Due to the additivity of these updates and the structure of the base learners (linear models, splines, tensor-products splines), an interpretable model is obtained. Due to its componentwise nature, CWB also comes with an inherent feature selection mechanism, similar to the Lasso \citep{Meinshausen.2007}. CWB can therefore be used for high-dimensional feature spaces ($n \ll p$ situations) and, when penalized properly, without a biased base learner selection towards more flexible terms~\citep{hofner2011framework}. Further advantages and extensions are given in Appendix~\ref{appendix:cwb}. 

Instead of using the vanilla CWB algorithm, we propose a novel stagewise procedure that has all the advantages of CWB but allows practitioners to better control overfitting and provides further insights into the modeled relationships.

\paragraph{Stagewise model fitting}

We define the final model $f$ to be a combination of three parts: $i$) univariate (linear + non-linear), $ii$) pairwise interactions, and $iii$) deep interactions, resulting in $f(\xb) = f_{\text{uni}}(\xb) + f_{\text{pint}}(\xb) + f_{\text{deep}}(\xb)$, fitted in three consecutive stages. The number of boosting iterations of each stage is dynamically selected by early stopping. Hence, if no risk improvement on a validation set is observed $\kappa$ consecutive times (our default: $\kappa = 2$), the fitting proceeds to the next stage.

In the first stage, we use CWB on all available features $\xb \in \mathbb{R}^p$ to explain as much information as possible through univariate (partial) effects $f_j$ defined on each single feature $x_j$ and aggregated in $f_{\text{uni}}(\xb) = \sum_{j=1}^p f_j(x_j)$. If $x_j$ is a numerical feature, $f_j$ is decomposed into a linear and a non-linear part, and CWB can choose to select either one or both. The linear part $b_{j,\text{lin}}$ consists of an intercept and a linear feature effect. The non-linear part $b_{j,\text{nlin}}$ describes the deviation from this linearity using a penalized B-spline effect \citep{Eilers.1996} centered around the linear effect. Centering the non-linear effect allows equal degrees-of-freedom to be defined for both base learners $b_{j,\text{lin}}$ and $b_{j,\text{nlin}}$, ensuring a fair selection between these two parts and an unbiased analysis of variance. Categorical features are included as linear effects using a dummy-encoding.

The second stage builds the model part $f_{\text{pint}}$ containing pairwise interactions $f_{ij}$ to model the interaction between features $x_i$ and $x_j$, $i\neq j$ in $\xb$. We initialize the model using the predictions $\hat{f}_{\text{uni}}(\xb)$ from the first stage as an offset and start with the corresponding pseudo residuals. These interactions are included using bivariate interactions of categorical variables, varying coefficient models \citep{Hastie.1993} for mixed categorical-numerical interactions, and penalized tensor-product splines~\citep{wood2017tensorsplines} for bivariate numerical interactions. For larger $p$, considering all possible pairwise interactions is infeasible. We use a filtering technique to include the $\psi\cdot100\%$ most frequently selected interactions $\mathcal{I}$ for $\psi\in(0,1]$ in a random forest (RF) selection step~\citep{breiman2001random}. The RF uses 500 trees and tree depth of two to allow the selection of only pairwise interactions. We then use CWB again to fit $f_{\text{pint}} = \textstyle\sum_{(i,j)\in\mathcal{I}} f_{ij}(x_i,x_j)$ and refine the set of possible interactions due to its selective nature. As for the univariate model, the degrees-of-freedom are set equal for each $f_{ij}$ to ensure a fair interaction selection. We note that the result of this stage is still interpretable, and bivariate interactions can, e.g., be visualized in a surface plot as shown in Section~\ref{sec:practicals}. 

A final third stage is used to explain the remaining variance left after the second stage. In contrast to the previous two stages, we use here a black-box model $f_{\text{deep}}(\xb) = \sum_{m=1}^{M_{\text{deep}}} T_m(\xb)$ including $M_{\text{deep}}$ deep trees $T_m$ as base learners. As for the second stage, the third stage starts with the pseudo residuals obtained after the second stage. This stage is able to represent deeper interactions and non-smooth feature effects. We can understand this stage as a measure of complexity needed to fit the given data and also as a measure of uncertainty about stages one and two. The smaller the fraction needed for stage three in the final model, the higher our confidence is with respect to the model's interpretation.

\subsection{The Bigger Picture}\label{subsec:autocboost-parts}

\begin{figure}[b]
    \centering
    \includegraphics[trim={0 0.3cm 0 1cm},width=0.9\textwidth]{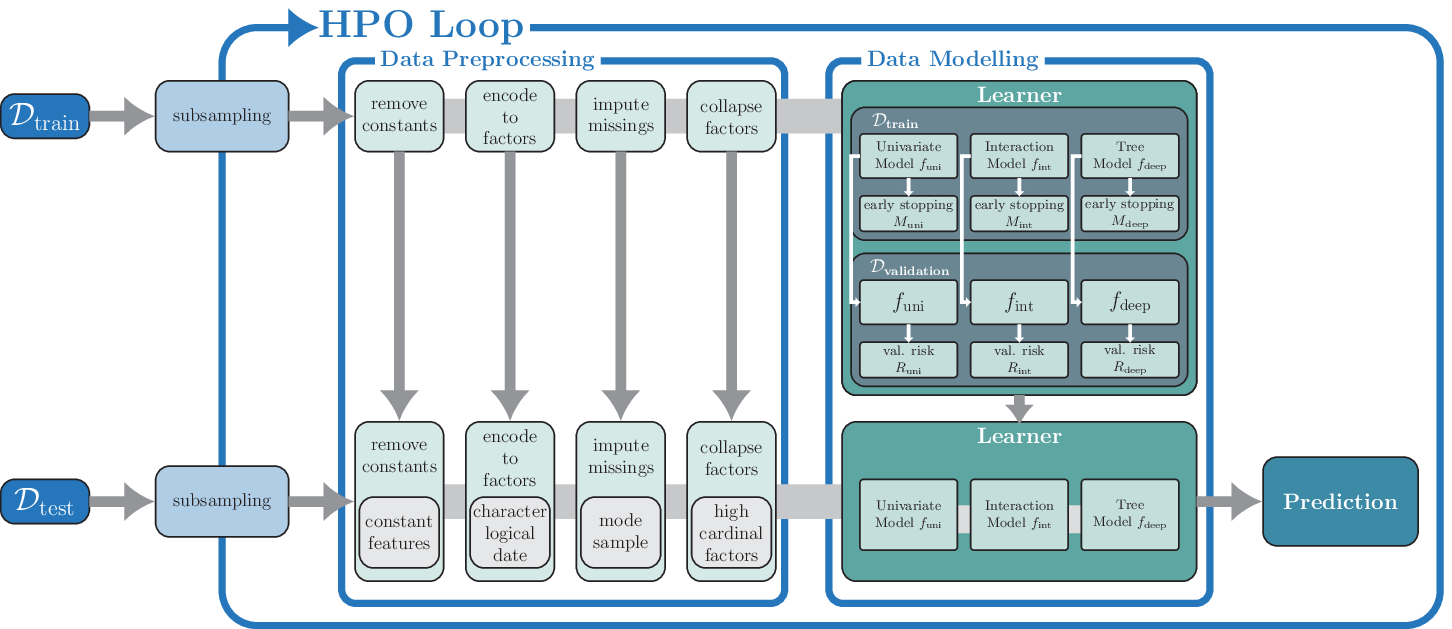}
    \caption{\texttt{autocompboost}'s standard ML pipeline.}
    \label{fig:acb_pipeline}
\end{figure}
Our framework consists of three blocks composed into an AutoML pipeline. These blocks are data preprocessing, data modelling, and HPO (illustrated in Fig.~\ref{fig:acb_pipeline} as blue frames). The first preprocessing block is automatically determined by the underlying task. This includes the removal of constant features, feature encoding, and collapsing levels of high cardinal factors. It also implements an automatic imputation of missing values (see Appendix~\ref{appendix:mv} for details). The second block contains the model fitting (Section~\ref{subsec:cboost}), while the third block is formed by HPO. We use Hyperband \citep{lisha2018hyperband} here, but the framework allows using any other tuning algorithm. Common applications of Hyperband are using the number of iterations as a multifidelity budget parameter. In our three-stage approach, we determine the number of boosting iterations in each stage using early stopping. Hence, instead of the boosting iterations, a subsampling rate is used as budget parameter. More specifically, multiple candidate models are fitted on a fraction of the complete data set based on the current subsampling rate and only the most promising candidates are used further for larger fractions. Due to the choice of the fitting algorithm, another advantage is the rather small HPO search space, consisting only of HPs $\nu \in [0.001, 0.5]$ and $\psi \in [0.01, 0.2]$.
\vspace{-0.1cm}

\subsection{Implementation}
The proposed framework is implemented in the \texttt{R} package \texttt{autocompboost} based on \texttt{compboost}~\citep{schalk2018compboost} and the \texttt{mlr3} ecosystem~\citep{mlr3}.
Model fitting is executed in \texttt{C++}, naturally allowing for fast and parallel computing.
The task-specific ML pipeline is defined via \texttt{mlr3pipelines}~\citep{binder2021pipelines}, while HPO is based on Hyperband \citep{lisha2018hyperband} using \texttt{mlr3hyperband}.

\vspace*{-0.1cm}
\section{Benchmark}\label{sec:benchmark}

To demonstrate that the proposed framework can keep up with performances of state-of-the-art AutoML frameworks, we run our system on the OpenML AutoML Benchmark~\citep{amlb2019}. The benchmark is open-source -- allowing for easy and fair comparisons -- and includes 39 data sets. For the comparison, each AutoML system is trained for 1 hour on each outer resampling fold (10 fold cross-validation).
More details can be found in Appendix~\ref{appendix:benchmark}. Note that the dataset portfolio only contains classification tasks, while our framework can also be used without any restrictions for other types of tasks such as regression with different loss functions and modelling count, functional, or survival data. On a selection of 29 datasets, we run four different \texttt{autocompboost} configurations -- with and without deep interactions, both with and without HPO. Additionally to the AutoML systems, \citet{amlb2019} provide results for a (tuned and non-tuned) random forest and a constant predictor. We added a \texttt{glmnet} model, tuned for 1 hour via random search to the comparison. The results are shown in Figure~\ref{fig:results} below and in detail in Table~\ref{tab:results} in Appendix~\ref{appendix:benchmark}. Results indicate that \texttt{autocompboost} is in most tasks very competitive to other existing systems.
\vspace{-0.1cm}
\begin{figure}[H]
    \centering
    \includegraphics[trim={0 0 0 0.7cm}, width=0.8\textwidth]{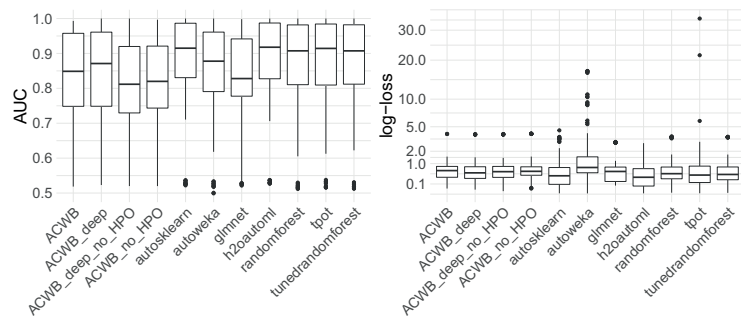}
    \caption{Benchmark results across all datasets for different AutoML systems (Auto-WEKA, TPOT, H2O-AutoML, auto-sklearn), RF (tuned and with default values) as an ML model with good out-of-the-box performance, a GLM with elastic net regularization as a comparable interpretable model, as well as four variants of our framework (ACWB). Left: boxplots of AUC performances for binary classification tasks. Right: log-loss results for multiclass classification tasks.}
    \label{fig:results}
\end{figure}

\vspace{-1cm}

\section{Interpreting the framework}\label{sec:practicals}

\texttt{autocompboost} automatically provides three important ways to interpret the final model: 1) the required complexity based on the decomposition of the model's three stages; 2) feature or variable importance (VIP); 3) the estimated partial effects $f_j$ and $f_{ij}$ of the first and second stage. For the sake of illustration, we use the Adult data set described in detail in Appendix~\ref{appendix:application}.

\paragraph{Required model complexity}

During training, the train and validation risk is logged for each stage. This allows inference about the required model complexity by calculating the percentage of explained risk per stage. The explained risk of the univariate stage can be further decomposed into risk reduction by linear and non-linear effects. In particular, the fraction of explained risk within stages one and two divided by the total explained risk describes the degree to which the interpretable model contributes to the final predictions.

\vspace*{-0.cm}
\begin{figure}[H]
    \centering
    \includegraphics[trim={0 0.4cm 0 1cm}, width=0.6\textwidth]{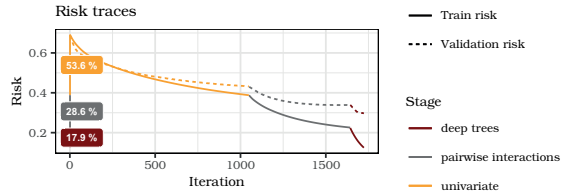}
    \caption{Explained risk per stage and iteration mapped to the percentage of explained risk per stage as an indicator for the required complexity. Here, 53.6 \% of the risk is explained by the $f_{\text{uni}}$, 28.6 \% by $f_{\text{pint}}$, and 17.9 \% using deep trees.}
    \label{fig:complexity}
\end{figure}
\vspace*{-0.8cm}
\paragraph{Variable importance}

Similar to the overall risk reduction, the VIP is the risk reduced per feature, for which we make use of the base learner structure (details are given in Appendix~\ref{appendix:vip}). The VIP allows the user to investigate which feature in the model is able to reduce the risk and to what extent. 

\paragraph{Explaining the model's decision-making}

\begin{figure}
    \centering
    \includegraphics[trim={0 0.8cm 0 1.6cm}, width=0.8\textwidth]{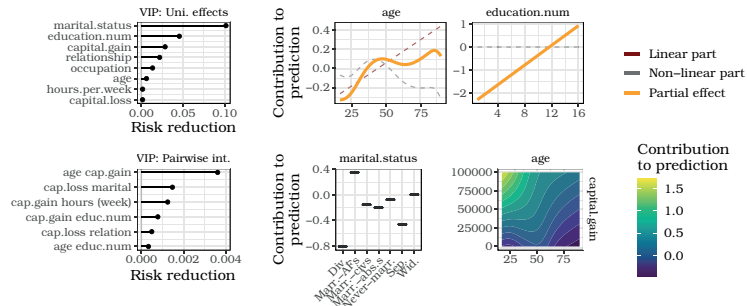}
    \caption{VIP (left) of stages one and two and Partial effects (right) of two numerical features with their decomposition into linear and non-linear effect (right, top), a categorical feature (right, bottom left), and an interaction (right, bottom right).}
    \label{fig:vip-pe}
\end{figure}

Our routine provides two ways to explain the model's decision-making. The first visualizes the partial effects and pairwise interactions (Figure~\ref{fig:vip-pe}). The second shows how each feature contributes to the prediction score for a new observation (Figure~\ref{fig:pred-decomp} in Appendix~\ref{appendix:application}).

\newpage

\section*{Acknowledgements}
This work has been partially supported by the German Federal Ministry of Education and Research (BMBF) under Grant No. 01IS18036A and by the Federal Ministry for Research and Technology (BMFT) grant FKZ: 01ZZ1804C (DIFUTURE, MII). The authors of this work take full responsibilities for its content.

\begingroup
\let\chapter\section

\renewcommand{\bibname}{References}

\bibliographystyle{chicago}
\bibliography{refs}

\endgroup

\newpage
\appendix

\section{Componentwise boosting}\label{appendix:cwb}

CWB~\citep{buhlmann2003boosting} uses gradient boosting~\citep{freund1996experiments} by sequentially adding one base learner $b_k$ out of a set of base learners $\mathcal{B} = \{b_k\ |\ k = 1, \dots, K\}$ to the model. The objective of boosting is to minimize the empirical risk $$\mathcal{R}_f(\D) = \sum_{(\xb,y)\in\D} L(y, f(\xb))$$ w.r.t. a prediction model $f$ with loss function $L$ and data $$\D = \Dset$$ containing the target vector $\yvec\in\R^n$ and features $$\xb_j=\xjvec\in\R^n.$$ The collection of features is denoted as observations $\xi = \xivec\in\R^p$. 

The prediction model $\fmh$ after $m$ boosting iterations is defined using an additive structure $$\fmh(\xb) = \fmdh(\xb) + \nu b_{k^{[m]}}(\xb|\thetabh^{[m]}),$$, where the best-performing base learner $b_{k^{[m]}}(\xb|\thetabh^{[m]})$ out of all base learners in $\mathcal{B}$ is added in each iteration $m$. To obtain an interpretable model, $b_k(\xb| \thetab_k)$ is parametrized by a structured additive model with parameter $\thetab_k\in\R^{d_k}$ and estimate $\hat{\thetab}^{[m]}_k$ in iteration $m$. The offset of the model $f^{[0]} = \argmin_{c\in\R}\mathcal{R}_c(\D)$ is found by choosing the constant $c$ that minimizes the risk. To find the best base learner $k^{[m]} \in \{1, \dots, K\}$ in iteration $m$, each base learner $b_k$ is fitted to the so-called pseudo residuals $\rmm$ with $$\rmi = -\fp{\Lxyi}{f(\xi)}\left.\vphantom{f_{(i}^{(i}}\right|_{f = \fmdh},$$  $i =1, \dots, n$. Therefore, the sum of squared errors $$\text{SSE}^{[m]}_k(\thetab_k) = \sum_i^n (b_k(\xi |\thetab_k) - \rmi)^2$$ is minimized to find the best parameter estimates $\thetabh^{[m]}_k = \argmin_{\thetab_k} \text{SSE}^{[m]}_k(\thetab_k)$ for all $b_k\in\mathcal{B}$. The index $k^{[m]} = \argmin_{k\in\{1, \dots, K\}} \text{SSE}^{[m]}_k(\thetabh^{[m]}_k)$ of the best model is then chosen by the smallest SSE. The parameter of the corresponding base learner of this iteration is defined as $\thetabh^{[m]} = \thetabh^{[m]}_{k^{[m]}}$.

A base learner $b_k$ selects the feature(s) of the input vector $\xb = (x_1, \dots, x_p)^\tran \in \R^p$ required for modelling, e.g., the univariate linear regression model or splines. This can be one feature $x_i$, but also two features $x_i$ and $x_j$ for tensor splines\footnote{It is also possible to model more complex dependencies with a base learner, e.g., when using tree base learners. In this paper with a focus on interpretability, we restrict ourselves to regression models that can be represented by linear feature effects (after a B-spline basis function evaluation).}. This allows estimating the partial feature effects as linear combination of feature (or its basis representation) and the effect $\thetabh_k$.

The iterative boosting process is performed until a predefined number of boosting iterations is reached. Another option is to use stopping mechanisms, such as early stopping, or using a time budget for a dynamic stopping of the fitting process. We denote the number of boosting iterations used to fit the model with $M\in\N$.

\paragraph{Parameter aggregation} 

After fitting CWB, the whole model is defined by the sequence of selected base learners with estimated parameters $\{\thetabh^{[1]}, \dots, \thetabh^{[M]}\}$. We restrict the base learner to be linear in these parameters $b_j(\xb|\thetab_j) = \xb^\tran \thetab_j$. Therefore, two base learners $b_{j^\ast}(\xb| \hat{\thetab}^{[m]})$ and $b_{j^\ast}(\xb| \hat{\thetab}^{[m^\prime]})$ of the same type $j^\ast$ with parameter vectors $\hat{\thetab}^{[m]}$ and $\hat{\thetab}^{[m^\prime]}$ can be aggregated to $b_{j^\ast}(\xb|\hat{\thetab}^{[m]}+\hat{\thetab}^{[m^\prime]})$. This is further used to obtain parameter estimates for all base learners by calculating $\thetabh_j = \nu \sum_{m=1}^M\sum_{k=1}^K\mathds{1}_{\{j=k^{[m]}\}}\thetabh^{[m]}$ and is key for an inherent partial effects estimation and interpretation. The partial effect of the $j$-th feature $f_j(\xb)$ is defined as aggregated base learner $b_j(\xb | \thetabh_j)$.

\paragraph{Extensions and applications} CWB allows optimization of arbitrary differentiable loss functions and can thus be used for, e.g., multiclass classification, interval or survival regression, and probabilistic forecasts. It is also possible to quantify the epistemic uncertainty of CWB \citep{rugamer2019inference}. Many other extensions of the CWB algorithm exist, such as CWB for functional data \citep{JSSv094i10}, boosting location, scale and shape models \citep{hofner2014gamboostlss}, or probing for sparse and fast variable selection \citep{thomas2017probing}. Because of its interpretability properties, CWB is used in medical research, e.g., for oral cancer prediction \citep{saintigny2011gene}, detection of synchronization in bioelectrical signals \citep{rugamer2016boosting}, or classifying pain syndromes \citep{liew2020classifyingneck}. 

\section{Univariate model}

The first stage of our fitting procedure uses only univariate models, and each of the $p$ features is included as separate base learner. Categorical features are included using a dummy encoding with a ridge penalty. Hence, if a categorical base learner is selected, all group parameters are updated at once. Numerical features are included by decomposing their effect into a linear and a non-linear effect. Hence, for each numerical feature $x_j$, two base learners $b_{j,\text{lin}}$ and $b_{j,\text{centered}}$ are defined. The parameter vector $\theta_{j,\text{lin}} = (\alpha_j, \beta_j)^\tran$ of $b_{j,\text{lin}}$ contains a feature-specific intercept $\alpha_j$ and a slope $\beta_j$. The centered base learner $b_{j,\text{centered}}$ uses a B-spline basis~\citep{eilers1996flexible}, where the parameter vector $\gamma_j\in\R^{p_j}$ contains the weights of the $p_j$ B-spline basis functions $B_1(x_j), \dots, B_{p_j}(x_j)$. The B-spline basis is then centered around the linear effect to subtract the linear part from the basis. The non-linear base learner is then estimated using penalized least squares with a P-spline penalty. The partial effect $f_j$ of the numerical feature $x_j$ is thus given as sum of the two univariate base learners $f_j = b_{j,\text{lin}} + b_{j,\text{centered}}$. 

\section{Variable importance} \label{appendix:vip}

The VIP of feature $j$ is defined as $$\text{VIP}_j = \sum_{m=1}^{M_{\text{uni}}} (\mathcal{R}_{\hat{f}^{[m-1]}_{\text{uni}}}(\D) - \mathcal{R}_{\hat{f}^{[m]}_{\text{uni}}}(\D))\mathds{1}(k^{[m]} = j).$$ The same formula is applied to calculate the $\text{VIP}_{ij}$ for interactions $(i,j)\in\mathcal{I}$ and $\hat{f}_{\text{pint}}$. For the Adult data, as shown in Figure~\ref{fig:vip-pe}, the two most important features during the univariate fitting stage are the marital status (\texttt{marital.status}) and a numeric representation of the education status (\texttt{education.num}), with the marital status reducing the risk twice as much as the education status. The dominating interaction is between age (\texttt{age}) and capital gain (\texttt{capital.gain}).

\section{The Bigger Picture} \label{appendix:mv}

Missing factors in our framework are by default imputed by their mode, while missing numeric values are sampled from all possible values with probabilities according to their empirical distribution. Hyperband, the algorithm used for HPO, can best be understood as repeated execution of the \emph{successive halving} procedure \citep{Jamieson2016NonstochasticBA}.

\section{Benchmark setup}\label{appendix:benchmark}
The original OpenML AutML benchmark consists of 39 datasets.
Small- and medium-sized datasets are trained for 1 and 4 hours respectively, large datasets for 4 and 8 hours. 
The benchmark was initially run on Amazon Web Services \textit{m5.2xlarge} (8 CPUs) instances inside a docker container.
For our benchmark, we focus on the small to medium datasets with details described in Table~\ref{tab:datasets}.

We use the infrastructure of the Leibniz Supercomputing Centre (LRZ), operated by the Bavarian Academy of Sciences and Humanities.
Using the \texttt{R}-package \texttt{batchtools} \citep{lang2017batchtools}, our batch jobs ran on 8 core Haswell-based CPUs and 16Gb memory. On larger datasets and, in particular, multiclass tasks, \texttt{autocompboost} trained longer than 1 hour due to a preliminary and rather inefficient implementation using a one-versus-rest fitting procedure.
Moreover, the benchmark results reveal that the implemented HPO method still offers potential for improvement.
\begin{table}[ht!]
    \tiny
    \centering
    \begin{tabular}{rlrrrrr}
ID & Name & \# Instances & \# Features  & \# Classes & \# Missings & \# Numeric Features\\
\midrule
3 & kr-vs-kp & 3196  & 37  & 2 & 0 & 0 \\
12 & mfeat-factors & 2000  & 217 & 10  & 0 & 216 \\
31 & credit-g & 1000  & 21  & 2 & 0 & 7 \\
53 & vehicle & 846 & 19  & 4 & 0 & 18  \\
3917 & kc1 & 2109  & 22  & 2 & 0 & 21  \\
3945 & KDDCup09\_appetency & 50000 & 231 & 2 & 8024152 & 192  \\
7592 & adult & 48842 & 15  & 2 & 6465  & 6 \\
9952 & phoneme & 5404  & 6 & 2 & 0 & 5  \\
9977 & nomao & 34465 & 119 & 2 & 0 & 89 \\
10101 & blood-transfusion-service-center & 748 & 5 & 2 & 0 & 4 \\
14965 & bank-marketing & 45211 & 17 & 2	& 0 & 7 \\ 
146195 & connect-4 & 67557 & 43  & 3 & 0 & 0    \\
146212 & shuttle & 58000 & 10  & 7 & 0 & 9    \\
146606 & higgs & 98050 & 29  & 2 & 9 & 28     \\
146818 & Australian  & 690 & 15  & 2 & 0 & 6    \\
146821 & car & 1728  & 7 & 4 & 0 & 0    \\
146822 & segment & 2310  & 20  & 7 & 0 & 19     \\
146825 & Fashion-MNIST & 70000 & 785 & 10  & 0 & 784    \\
167120 & numerai28.6  & 96320 & 22  & 2 & 0 & 21     \\
168329 & helena & 65196 & 28  & 100 & 0 & 27     \\
168330 & jannis & 83733 & 55  & 4 & 0 & 54     \\
168331 & volkert & 58310 & 181 & 10  & 0 & 180    \\
168335 & MiniBooNE & 130064  & 51  & 2 & 0 & 50     \\
168337 & guillermo & 20000 & 4297  & 2 & 0 & 4296     \\
168338 & riccardo & 20000 & 4297  & 2 & 0 & 4296     \\
168908 & christine & 5418  & 1637  & 2 & 0 & 1599     \\
168909 & dilbert & 10000 & 2001  & 5 & 0 & 2000     \\
168911 & jasmine & 2984  & 145 & 2 & 0 & 8    \\
168912 & sylvine & 5124  & 21  & 2 & 0 & 20     \\
\midrule
    \end{tabular}
    \caption{30 small to medium sized datasets of the OpenML AutoML Benchmark.}
    \label{tab:datasets}
\end{table}

\begin{landscape}
\begin{table}
    \tiny
    \centering
\begin{tabular}{l|lrrrrrrrrrrr}
Metric & Dataset & ACWB & ACWB\_deep & ACWB\_deep\_no\_HPO & ACWB\_no\_HPO & autosklearn & autoweka & glmnet & h2oautoml & randomforest & tpot & tunedrandomforest\\
\midrule
AUC & Australian & 0.926 & 0.911 & 0.922 & 0.927 & 0.935 & 0.929 & 0.929 & \textbf{0.940} & 0.937 & 0.932 & 0.934\\
AUC & KDDCup09\_appetency & 0.705 & 0.714 & 0.724 & 0.734 & 0.834 & 0.808 & 0.792 & \textbf{0.830} & 0.786 & 0.824 & 0.786\\
AUC & MiniBooNE & 0.963 & 0.965 & 0.958 & 0.955 & 0.985 & 0.961 & 0.936 & \textbf{0.987} & 0.982 & 0.981 & 0.982\\
AUC & adult & 0.900 & 0.904 & 0.911 & 0.911 & \textbf{0.930} & 0.908 & 0.904 & 0.926 & 0.909 & 0.927 & 0.909\\
AUC & bank-marketing & 0.854 & 0.898 & 0.897 & 0.855 & \textbf{0.937} & 0.827 & 0.909 & \textbf{0.937} & 0.931 & 0.934 & 0.931\\
AUC & blood-transfusion & 0.755 & 0.750 & 0.749 & 0.725 & \textbf{0.757} & 0.741 & 0.754 & 0.756 & 0.686 & 0.724 & 0.689\\
AUC & christine & 0.788 & 0.791 & 0.802 & 0.793 & \textbf{0.830} & 0.802 & 0.800 & 0.826 & 0.806 & 0.813 & 0.810\\
AUC & credit-g & 0.763 & 0.755 & 0.760 & 0.768 & 0.783 & 0.753 & 0.786 & 0.789 & 0.795 & 0.786 & \textbf{0.796}\\
AUC & guillermo & 0.786 & 0.780 & 0.752 & 0.753 & 0.901 & 0.878 & 0.771 & \textbf{0.910} & 0.903 & 0.819 & 0.903\\
AUC & higgs & 0.745 & 0.746 & 0.761 & 0.762 & 0.793 & 0.677 & 0.680 & \textbf{0.814} & 0.803 & 0.802 & 0.803\\
AUC & jasmine & 0.841 & 0.848 & 0.849 & 0.843 & 0.884 & 0.861 & 0.849 & 0.888 & 0.888 & 0.885 & \textbf{0.889}\\
AUC & kc1 & 0.790 & 0.791 & 0.793 & 0.795 & 0.840 & 0.814 & 0.799 & 0.836 & 0.836 & 0.841 & \textbf{0.842}\\
AUC & kr-vs-kp & 0.976 & 0.999 & 0.998 & 0.994 & \textbf{1.000} & 0.976 & 0.995 & \textbf{1.000} & 0.999 & \textbf{1.000} & \textbf{1.000}\\
AUC & nomao & 0.985 & 0.984 & 0.988 & 0.989 & \textbf{0.996} & 0.984 & 0.988 & \textbf{0.996} & 0.995 & 0.995 & 0.995\\
AUC & numerai28.6 & 0.528 & 0.528 & 0.527 & 0.526 & 0.529 & 0.520 & 0.529 & \textbf{0.532} & 0.520 & 0.525 & 0.521\\
AUC & phoneme & 0.906 & 0.909 & 0.920 & 0.919 & 0.963 & 0.957 & 0.813 & 0.968 & 0.965 & \textbf{0.969} & 0.966\\
AUC & riccardo & 0.949 & 0.956 & 0.756 & 0.847 & \textbf{1.000} &  & 0.996 & \textbf{1.000} & 0.999 & 0.992 & \textbf{1.000}\\
AUC & sylvine & 0.972 & 0.973 & 0.977 & 0.974 & 0.990 & 0.975 & 0.966 & 0.990 & 0.983 & \textbf{0.992} & 0.984\\
\midrule
log-loss & Fashion-MNIST & 0.608 & 0.589 & 0.613 & 0.615 & 0.354 & 0.581 & 0.407 & \textbf{0.294} & 0.361 & 0.651 & 0.362\\
log-loss & car & 0.553 & 0.200 & 0.312 & 0.427 & 0.010 & 0.243 & 0.161 & 0.002 & 0.144 & \textbf{0.000} & 0.047\\
log-loss & connect-4 & 0.837 & 0.787 & 0.812 & 0.793 & 0.426 & 0.741 & 0.606 & \textbf{0.345} & 0.495 & 0.400 & 0.478\\
log-loss & dilbert & 0.377 & 0.283 & 0.424 & 0.433 & 0.097 & 584.504 & 0.146 & \textbf{0.052} & 0.328 & 0.217 & 0.329\\
log-loss & helena & 3.977 & 3.921 & 3.925 & 3.998 & 3.447 & 14.102 & 2.922 & \textbf{2.800} & 3.550 & 3.245 & 3.559\\
log-loss & jannis & 0.870 & 0.871 & 0.854 & 0.857 & 0.705 & 1077.099 & 0.832 & \textbf{0.681} & 0.728 & 0.732 & 0.729\\
log-loss & mfeat-factors & 0.267 & 0.245 & 0.152 & 0.173 & \textbf{0.099} & 0.627 & 0.106 & 0.105 & 0.234 & 0.138 & 0.201\\
log-loss & segment & 0.392 & 0.307 & 0.232 & 0.273 & 0.060 & 0.501 & 0.325 & \textbf{0.047} & 0.084 & 0.052 & 0.069\\
log-loss & shuttle & 0.087 & 0.028 & 0.024 & 0.059 & 0.001 & 0.015 & 0.118 & \textbf{0.000} & 0.001 & 3.444 & 0.001\\
log-loss & vehicle & 0.637 & 0.583 & 0.546 & 0.546 & 0.395 & 2.105 & 0.423 & \textbf{0.353} & 0.497 & 0.414 & 0.486\\
log-loss & volkert & 1.499 & 1.476 & 1.411 & 1.471 & 0.945 & 1.110 & 1.177 & \textbf{0.821} & 0.980 & 1.011 & 0.979\\
\midrule
\end{tabular}

    \caption{Results of the benchmark experiment based on 29 datasets (see Table~\ref{tab:datasets}). Results are presented as mean scores of a 10-fold crossvalidation. Best scores are presented as bold numbers. For the binary and multiclass classification tasks, the results are given by the AUC and the log-loss, respectively.}
    \label{tab:results}
\end{table}
\end{landscape}

\section{Required model complexity}

During model training of \texttt{autocompboost} for the three described stages, we denote the number of boosting iterations of each stage with $M_{\text{uni}}$, $M_{\text{pint}}$, and $M_{\text{deep}}$. Furthermore, the empirical risk of the intercept model $\mathcal{R}_0 = \mathcal{R}_{\fh^{[0]}}(\D)$, the univariate model $\mathcal{R}_{\text{uni}} = \mathcal{R}_{\fh_{\text{uni}}^{[M_{\text{uni}}]}}(\D)$, the pairwise interaction model $\mathcal{R}_{\text{pint}} = \mathcal{R}_{\fh_{\text{pint}}^{[M_{\text{pint}}]}}(\D)$, and the deep interaction model $\mathcal{R}_{\text{deep}} = \mathcal{R}_{\fh_{\text{deep}}^{[M_{\text{deep}}]}}(\D)$ is logged. 
We define the fraction of explained risk per stage with
\begin{align*}
    \rho_{\text{uni}} &= (\mathcal{R}_{0} - \mathcal{R}_{\text{uni}}) / \delta_\mathcal{R} \\
    \rho_{\text{pint}} &= (\mathcal{R}_{int} - \mathcal{R}_{\text{pint}}) / \delta_\mathcal{R} \\
    \rho_{\text{deep}} &= (\mathcal{R}_{pint} - \mathcal{R}_{\text{deep}}) / \delta_\mathcal{R}
\end{align*}
with $\delta_{\mathcal{R}} = \mathcal{R}_0 - \mathcal{R}_{\text{deep}}$. The value of $\rho$ indicates how much of the overall explained risk is explained by stage the respective stage. Hence, $\rho$ is an indicator of how much complexity of the model is required to obtain different levels of prediction accuracy. 

\section{Application}\label{appendix:application}

\paragraph{Adult data set}

The Adult data was collected by the Census bureau. The binary classification task is to predict whether an adult earns more than \$50,000. The given features are, for example, education, hours of work per week, or marital status. The predicted scores $\hat{f}\in\R$ are mapped via the logistic function $s(\hat{f}) = (1 + \exp(-\hat{f}))^{-1} \in (0,1)$ to $(0,1)$, which can be interpreted as probabilities for the positive class of earning more than \$50,000. Hence, a predicted partial effect greater than zero favors the prediction of the positive class.

\paragraph{Required model complexity}

Figure~\ref{fig:complexity} demonstrates that, for the Adult data, 53.6 \% of the risk is already explained by the univariate model and 28.6 \% by the pairwise interactions. Including deep trees into the model accounts for another 17.9 \% explained risk.

\paragraph{Explaining the models decision making}

Figure~\ref{fig:vip-pe} (middle) shows how partial effects of both numerical and categorical features can be visualized. In our example, the effect of the feature \texttt{age} shows both linear and non-linear effects. The second most important feature \texttt{education.num} only selects the linear component.

Pairwise interactions $f_{ij}$ are visualized by plotting the effect surface in the two feature dimensions. Figure~\ref{fig:vip-pe} demonstrates this, showing that especially younger adults with large capital gain likely have earnings greater than \$50,000 per year.

If the third stage has a significant contribution for explaining the final model fit, interpretation based on the first two stages must be performed with caution due to two reasons. First, the major part of the model's prediction stems from the black box model, and interpreting the structured partial effects alone is likely to be misleading. Second, a potential overlap in the hypothesis space between the interpretable stages and the black box stage might yield to an identifiability issue~\citep[see, e.g.,][]{ruegamer2021sddr}.

\paragraph{Prediction decomposition}

\begin{figure}
  \centering
  \includegraphics[width=\textwidth]{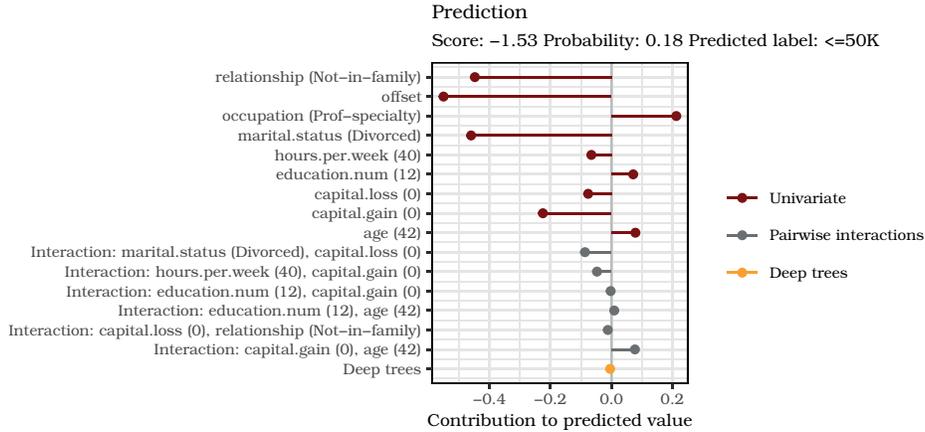}
  \caption{Decomposition of how a new predicted score is calculated. }
  \label{fig:pred-decomp}
\end{figure}

\texttt{autocompboost} also allows the user to better understand the system's decision-making process when a new observation $\xb_0$ is used to predict the score $\hat{f}(\tilde{\xb})$. In this case, it can visualize the contribution of all univariate effects by calculating $f_j(\tilde{x}_j)$, pairwise interactions with $f_{ij}(\tilde{x}_i, \tilde{x}_j)$, and the black box part $f_{\text{deep}}(\tilde{\xb})$. Figure~\ref{fig:pred-decomp} shows this decomposition. The contribution of the black box part is summarized as \enquote{deep trees} contribution. 

\paragraph{User-friendly interface} 

To further simplify the use of \texttt{autocompboost}, the previous explained techniques can be interactively visualized in a dashboard. Thereby, the user automatically obtains a pre-selection of important features, effects, and corresponding visualizations by only a few clicks. It is also possible to move the whole \texttt{autocompboost} pipeline into the dashboard, i.e., perform 1) task-creation, 2) data modelling and 3) interpretation in order to make interpretable ML models available to experienced ML users as well as a larger group of non-ML experts.

\end{document}